\pdfoutput=1
\documentclass[11pt]{article}
\IfFileExists{sections/introduction.tex}{\gdef\paperroot{}}{\gdef\paperroot{Arxiv/}}

\usepackage[T1]{fontenc}
\usepackage[utf8]{inputenc}
\usepackage{lmodern}
\usepackage[a4paper,margin=1in]{geometry}
\usepackage{microtype}
\usepackage{amsmath,amssymb}
\usepackage{graphicx}
\usepackage{booktabs}
\usepackage{multirow}
\usepackage{xcolor}
\usepackage{listings}
\usepackage[numbers,sort&compress]{natbib}
\usepackage{xurl}
\usepackage[colorlinks=true,linkcolor=blue!50!black,citecolor=blue!50!black,urlcolor=blue!50!black]{hyperref}
\usepackage[capitalise,noabbrev]{cleveref}
\usepackage{tikz}
\usetikzlibrary{arrows.meta,calc,decorations.pathreplacing,fit}
\usepackage{fontawesome5}
\usepackage{authblk}

\graphicspath{{\paperroot figures/}}
\newcommand{\ej}{EarthJev}

\newcommand{\todo}[1]{\textcolor{red}{[TODO: #1]}}

\title{\ej: Calibrated, Non-Generative Answers to Natural-Language Questions about Sentinel-2 Land Cover}

\newcommand{\orgdiv}[1]{#1}
\newcommand{\orgname}[1]{#1}
\newcommand{\orgaddress}[1]{#1}
\newcommand{\street}[1]{#1}
\newcommand{\city}[1]{#1}
\newcommand{\postcode}[1]{#1}
\newcommand{\state}[1]{#1}

\author{Domenico Benfenati}
\author{Antonio Maria Rinaldi}
\author{Cristian Tommasino}
\affil{\orgdiv{Department of Electrical Engineering and Information Technology (DIETI)}, \orgname{University of Naples Federico II}, \orgaddress{\street{Via Claudio, 21}, \city{Naples}, \postcode{80125}, \state{Italy}}}
\date{}

\begin{document}
\maketitle

\begin{abstract}
Vision-language models for remote sensing answer questions about satellite images by generating text: every answer costs a decoding loop, has to be parsed back to an option, and comes without a usable probability. We present \ej, a 0.8B-parameter model that answers typed questions about multispectral Sentinel-2 patches without generating any token. All 12 bands enter the vision encoder of Qwen3.5-0.8B through a widened input layer initialised from the RGB kernels, and, following the open OneJev recipe, the answer is read from the logits of the option letters at a single position and trained with cross-entropy plus Brier score. Under a pre-specified analysis plan, we compare it with the same backbone fine-tuned to generate the answer and with three multimodal language models fine-tuned with LoRA, all trained on the same 70{,}890 questions from 5\% of BigEarthNet v2.0 and timed in one job on the same GPU; VQA models that classify over a fixed answer vocabulary are reported as a reference for the cost of answering only the questions seen in training. \ej{} answers a question in 19.9\,ms, 4.3 to 8.7 times faster than the multimodal language models and 6.8 times faster than its own backbone used generatively; when all the questions about an image are asked together, it answers them in 42.8\,ms, 3.7 to 10.1 times faster than the batched generative models. On 36{,}028 test questions it reaches 86.0\% accuracy (three seeds), 0.6 and 1.2 points below the two 2B multimodal models and 3.5 points below the remote-sensing EarthDial-4B, with answers that are valid by construction and calibrated probabilities (ECE 0.006). Replacing the nine non-RGB bands with their mean costs 4.5 points, so the model uses the multispectral input, and in an ablation training the whole vision encoder raises accuracy to 88.4\% at the same cost. Code and model: \url{https://github.com/CosmoIknosLab/EarthJev}.
\end{abstract}

\section{Introduction}
\label{sec:intro}

Most questions that users ask about a satellite image have a closed set of admissible answers. \emph{Is there any forest in this patch? Which land-cover class dominates? Is there more arable land or more pasture? In which quadrant is the urban fabric?} Vision-language models (VLMs) for remote sensing~\citep{geochat,lhrsbot,earthdial,earthmind} answer such questions by generating free text, as they would answer an open-ended request. Using these answers in a decision has three costs. Every question costs an autoregressive decoding loop, whose length grows with the answer. The text has to be mapped back to one of the admissible options, and nothing guarantees that it matches one. And the model gives no probability for its answer, unless one is reconstructed from the token probabilities of the generated text.

A recent line of work replaces generation with a single typed decision. TypeSafe's \emph{System One} interface~\citep{typesafe} takes a state (text or JSON) and a set of typed questions---yes/no, one of $K$ options, or a score on an ordered scale---and returns a probability for every option instead of text. OneJev~\citep{onejev} is an open-source multimodal implementation of this interface: a Qwen vision-language model is fine-tuned to place the answer in the logits of the option letters at the last input position, with a proper scoring objective, so that a forward pass answers a question and the questions about the same state share its encoding. OneJev contains no remote-sensing data, and like every general VLM it reads three-channel images, while a Sentinel-2 patch has twelve spectral bands.

We ask whether this recipe gives a fast and reliable question-answering model for multispectral satellite images, and how it compares, under the same training data, with the models that do the same task. We make four contributions.

\begin{figure}[!t]
  \centering
  \resizebox{\linewidth}{!}{\input{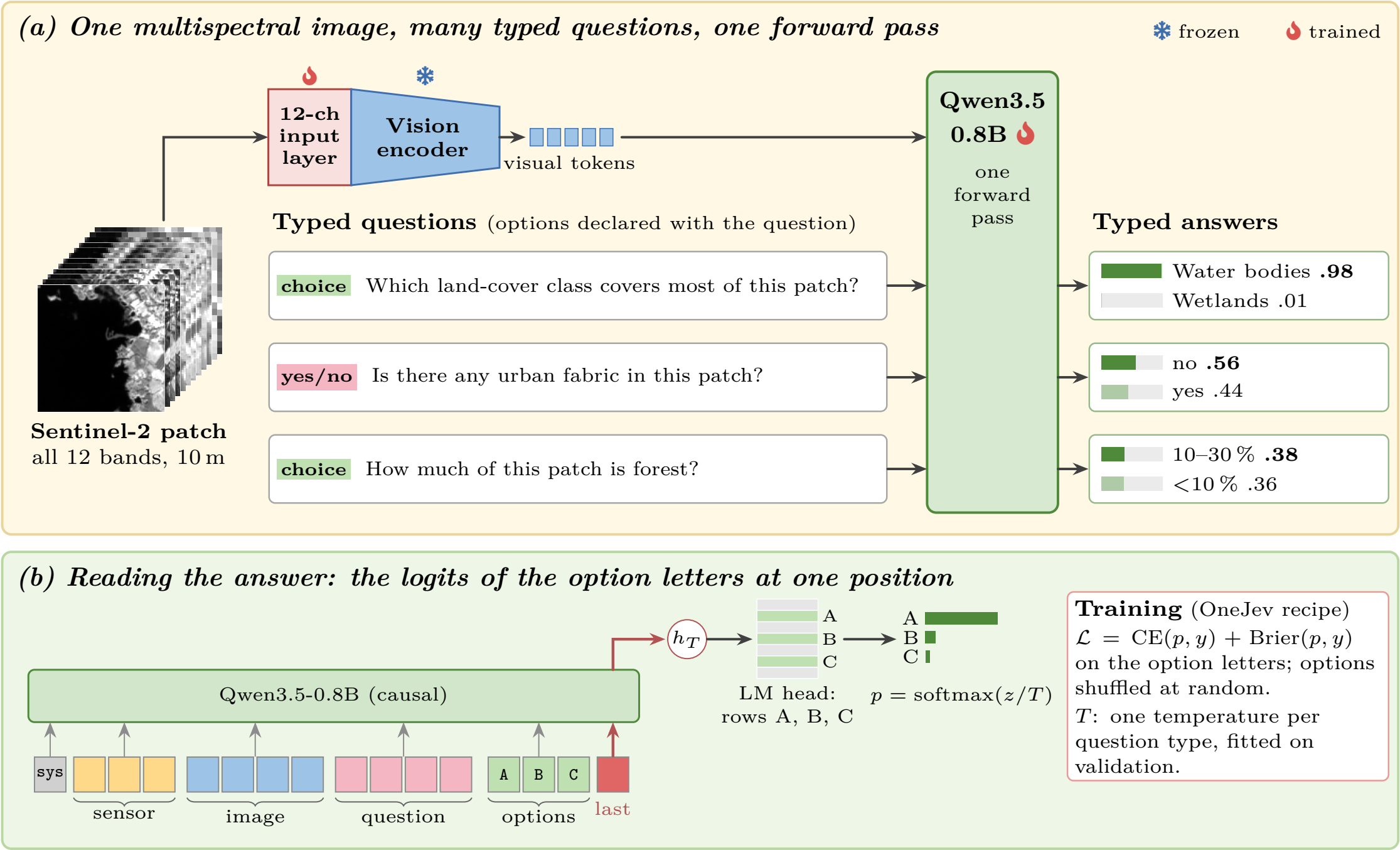}}
  \caption{\ej{} framework. \textbf{(a)}~All 12 bands of a Sentinel-2 patch enter the vision encoder of Qwen3.5-0.8B through a widened input layer, the only trained part of the encoder. Each typed question declares its options; in one forward pass the model returns a probability for every option of every question, and no other model's output enters its input. The probabilities are the outputs of the released model on a test patch (two most probable options): the first and third answers are right; urban fabric covers 11\% of the patch, so the second answer is wrong, with a probability close to 0.5. \textbf{(b)}~The answer is read from the logits of the option letters at the last position, so no token is generated; the model is trained with cross-entropy plus Brier score on these letters, and one temperature per question type is fitted on validation.}
  \label{fig:framework}
\end{figure}

\begin{itemize}
\item \textbf{\ej}, a 0.8B-parameter model that answers typed natural-language questions about Sentinel-2 patches from all 12 bands, with a probability for every option and without generating text (\cref{fig:framework}). The input layer of the vision encoder is widened to 12 channels so that, before training, the model computes exactly what the RGB model computes; this layer is the only change to the OneJev recipe (\cref{sec:method}).
\item \textbf{A typed CORINE benchmark} generated from the reference masks of BigEarthNet v2.0~\citep{reben}: 36{,}028 test questions of five types on 5{,}991 test patches, rendered a second time with templates held out from training (\cref{sec:data}).
\item \textbf{A controlled comparison under a pre-specified analysis plan}: the same backbone fine-tuned generatively and three multimodal language models fine-tuned with LoRA, all trained on the same questions with three seeds, tested on the CORINE benchmark and on two benchmarks written by other authors, and timed in one job on one GPU. VQA models with CNN, Transformer and Mamba image encoders, which classify over a fixed answer vocabulary, are trained on the same data as a reference for the cost of a model that answers only the questions it was trained on (\cref{sec:experiments}).
\item \textbf{Findings.} \ej{} is 4.3 to 8.7 times faster than the multimodal language models per question (3.7 to 10.1 times per image) and 6.8 times faster than its own backbone used generatively, at 0.6--1.2 points of accuracy from the 2B models, with calibrated probabilities; it uses the extra bands (4.5 points). It is not as accurate as the 4B remote-sensing model, and the frozen vision encoder of the recipe is its main limit (\cref{sec:ablations}).
\end{itemize}

Our main finding is that reading the answer instead of generating it makes a multispectral VLM several times faster at little cost in accuracy. With the same backbone, the same images and the same data, the one-pass read-out is 6.8 times faster than generation at 0.8 points of accuracy; its advantages are speed, answers that are valid by construction---which matters as soon as the options are new: the generative model answers outside the options for 45.9\% of the multiple-choice questions of a benchmark written by others---and calibrated probabilities obtained without decoding.

\section{Related work}
\label{sec:related}

\paragraph{Visual question answering in remote sensing.}
RSVQA~\citep{rsvqa} introduced visual question answering on remote-sensing images with questions generated automatically from geographic data. RSVQAxBEN~\citep{rsvqaxben} scaled the approach to BigEarthNet~\citep{bigearthnet}, generating presence and land-cover questions about Sentinel-2 patches from their labels. Models designed for these benchmarks combine an image encoder (a CNN in RSVQA, transformers in later work) with a question encoder and classify over a fixed answer vocabulary; examples are the multi-modal fusion transformer VBFusion~\citep{vbfusion} and the lightweight LiT-4-RSVQA~\citep{lit4rsvqa}, which is the reference model on RSVQAxBEN. They are trained on the benchmark's question distribution and do not accept arbitrary questions or option sets.

\paragraph{Remote-sensing vision-language models.}
GeoChat~\citep{geochat} adapts LLaVA-style instruction tuning to remote sensing; LHRS-Bot~\citep{lhrsbot} uses volunteered geographic information to align images and text; EarthDial~\citep{earthdial} is trained on 11 million instruction pairs covering RGB, SAR and multispectral imagery; EarthMind~\citep{earthmind} fuses data from several sensors. These models answer in free text. BigEarthNet.txt~\citep{bentxt} pairs 464{,}044 Sentinel-1/Sentinel-2 images with 9.6 million text annotations and provides a manually verified benchmark split. Its authors report that general and remote-sensing VLMs struggle with complex land-cover questions and that fine-tuning on the dataset helps; we use its benchmark split as a transfer test, without training on BigEarthNet.txt questions.

\paragraph{Discriminative decisions with multimodal language models.}
TypeSafe's System One interface~\citep{typesafe} returns typed, probabilistic values instead of text; we rely only on its public documentation for the interface and make no claim about how TypeSafe's models are built. OneJev~\citep{onejev} is an open-source implementation for multimodal inputs: Qwen vision-language models are fully fine-tuned so that the answer is read from the logits of the option letters at the last position, with a cross-entropy plus Brier loss, option permutation and per-type temperature scaling. Its training data contain no remote-sensing source. Decision models of this kind also exist outside remote sensing, for medical imaging~\citep{omnimedjev} and histopathology~\citep{pathjev}. RSJEV~\citep{rsjev}, developed concurrently with this work, introduces a one-pass decision module for remote-sensing scene classification and reports strong accuracy--efficiency trade-offs on UC Merced, AID and NWPU-RESISC45 with a 0.8B model. Its experimental plan compares the one-pass model with CNN, Transformer and Mamba classifiers and with general and remote-sensing MLLMs fine-tuned with LoRA; we follow the same plan. \ej{} differs in the task (natural-language questions of several types about medium-resolution Sentinel-2 patches, rather than scene classification over a fixed label set), in the input (all 12 Sentinel-2 bands through a widened input layer of the vision encoder, rather than an RGB image), and in the evaluation of calibration.

\paragraph{Calibration.}
Modern neural networks are often over-confident; temperature scaling is a simple and effective post-hoc fix~\citep{guo2017}. The Brier score~\citep{brier1950} and the log-loss are strictly proper scoring rules~\citep{gneiting2007}: they are minimised in expectation by the true probabilities, which is why OneJev, and therefore \ej, trains on both.

\section{Method}
\label{sec:method}

\subsection{Typed questions and single-position readout}
\label{sec:readout}

A request to \ej{} consists of a \emph{state} and one or more \emph{typed questions}. The state is a JSON object with a fixed description of the sensor (product level, bands, ground sample distance, patch size) and a placeholder for the image (\cref{lst:state} in \cref{app:state}); it contains nothing produced by another model. We use two of the three OneJev question types:
\begin{itemize}
\item \texttt{noul}: a statement or yes/no question; the output is $P(\text{yes})$;
\item \texttt{choice}: a question with $K\le255$ named options; the output is a distribution over the options and a confidence $(\max_k p_k - 1/K)/(1-1/K)$.
\end{itemize}
The third type, an ordered score on 2--10 levels, is available in the model but not used in our questions: quantities such as class shares are asked as \texttt{choice} questions over intervals.

Following OneJev~\citep{onejev}, the prompt lists the options under letters (A, B, C, \dots) after the state and the question, and a fixed system prompt asks for the letter only. The model never generates: we take the logits $z_k$ of the $K$ letter tokens at the last input position and answer with $p=\operatorname{softmax}(z/T_{\tau,K})$, where $T_{\tau,K}$ is a temperature fitted per question type $\tau$ and number of options $K$ on validation data (\cref{sec:training}). Only the output-head rows of the $K$ letters are computed. Questions about the same image share its prefix: the inference engine of OneJev encodes the image and the state once and evaluates all question suffixes in one batched forward pass.

The answer returned to a user is a deterministic function of $p$, e.g.\ ``Yes ($p=0.95$)'' or the most probable option with its probability. No text is generated, so an answer cannot fall outside the declared options and its probability is the model's own.

\subsection{Twelve Sentinel-2 bands in a vision-language model}
\label{sec:msi}

The vision encoder of Qwen3.5~\citep{qwen35} starts with a 3D convolution that maps each $16\times16$-pixel patch of two (repeated) frames of an RGB image to a token embedding; the number of input channels is a configuration parameter. We widen this layer from 3 to 12 channels, one per Sentinel-2 band (B01--B12 with B8A; B10 is not part of L2A products). The kernels of the R, G and B channels are copied onto B04, B03 and B02, the kernels of the nine other bands start at zero, and the bias is unchanged. Before training, the widened model therefore computes exactly what the RGB model computes on the same red, green and blue values; we verified that its logits are identical on 256 test questions.

\paragraph{Pixels.} The 20\,m and 60\,m bands are resampled to 10\,m by pixel repetition (as distributed in reBEN~\citep{reben}), so a $120\times120$ patch is a $12\times120\times120$ array. B04, B03 and B02 go through the fixed linear stretch of a true-colour composite (2nd to 98th percentile of the training split mapped to $[0,1]$) and then through the normalisation of the Qwen processor, as an RGB image would; the other nine bands are standardised with their mean and standard deviation on the training split. The array is then resized to $256\times256$ by the processor's own bicubic resize and cut into patches by its own patching code, so the number of visual tokens is that of an RGB image of the same size: 64 after the $2\times2$ merge.

\paragraph{What is trained.} OneJev keeps the vision encoder frozen. We train its widened input layer, which the new bands need, and keep every other weight of the encoder frozen. The input layer uses the same learning rate as the rest of the model: in a pilot on validation data, with a selection rule fixed before running it, $5\times10^{-6}$ gave a lower validation NLL than $10^{-4}$ and $10^{-3}$. This is the only change to the OneJev recipe.

\subsection{Model and training}
\label{sec:training}

\ej{} starts from Qwen3.5-0.8B~\citep{qwen35} and follows the official 0.8B OneJev configuration without changes to the hyperparameters: full fine-tuning of the language model in bfloat16, learning rate $5\times10^{-6}$ with 3\% warm-up, no weight decay, gradient clipping at 1.0, one epoch, token-budget batches (16{,}384 tokens, 64 rows per optimiser step). Only the last position is supervised. The loss is the cross-entropy between the softmax over the $K$ option letters and the target distribution, plus the Brier score of the same distribution with weight 1. During training the order of the options is permuted at random for every example, so that the model cannot rely on positions, and the JSON state is serialised compactly with probability 0.5 instead of with the indented form used at inference.

After training, we fit one temperature per (question type, $K$) by minimising the negative log-likelihood on held-out validation questions, as in OneJev. Training on the 70{,}890 questions of \cref{sec:data} takes about 21 minutes on two H100 GPUs.

\section{Data and benchmarks}
\label{sec:data}

\subsection{Images}
\label{sec:images}

We use BigEarthNet v2.0 (reBEN)~\citep{reben}: $120\times120$ Sentinel-2 L2A patches with 12 bands resampled to 10\,m and dense CORINE level-3 reference masks, in the official geographic train/validation/test split (CDLA-Permissive-1.0). To keep storage and compute small we draw a 5\% subset, stratified by split and country: 24{,}000 patches (11{,}893 train, 6{,}116 validation, 5{,}991 test), 2.6\,GB in Zarr. Every model sees the patch through the input described in \cref{sec:msi} (12 bands) or, for models that accept only three channels, as a true-colour composite (B4, B3, B2) with the same fixed stretch. The level-1 composition of the test patches is 45.1\% forest and semi-natural areas, 38.9\% agricultural areas, 12.9\% water bodies, 2.4\% artificial surfaces and 0.7\% wetlands.

\subsection{Typed CORINE questions}
\label{sec:corine-questions}

We generate questions from the CORINE reference mask of each patch at levels 1 (5 classes) and 2 (15 classes). Every answer is computed from the mask, and cases near a decision boundary are skipped instead of labelled:
\begin{itemize}
\item \textbf{dominant} (\texttt{choice}, 5 or 15 options): the class with the largest area, asked only if it exceeds the second by at least 5 points;
\item \textbf{presence} (\texttt{noul}): ``yes'' if the class covers at least 5\% of the patch, ``no'' if it is absent; one positive and one negative question per patch when available;
\item \textbf{share} (\texttt{choice}, 5 options): the share of a level-1 class in the intervals 0\%, $<$10\%, 10--30\%, 30--60\%, $>$60\%, skipped within 2 points of an interval edge and for shares between 0 and 2\%; with probability 0.5 the class is drawn among those present, otherwise among all classes;
\item \textbf{location} (\texttt{choice}, 4 options): the quadrant that holds most of a class, asked if the class covers at least 10\% of the patch and at least half of it lies in one quadrant;
\item \textbf{compare} (\texttt{choice}, 2 options): which of two classes covers more area, asked if their shares differ by at least 15 points and the larger is at least 10\%.
\end{itemize}
Questions are rendered in English from templates split into \emph{seen} templates, used for training and validation, and \emph{unseen} templates: the test set is written twice with the same question instances, once with each template set, to measure generalisation to new phrasings. The splits contain 70{,}890 training, 36{,}698 validation and 36{,}028 test questions. Answering the test questions from the reference masks with the same rules gives 100\% accuracy, which checks the generator. \Cref{tab:templates} in \cref{app:templates} lists the templates.

\subsection{Transfer tests}
\label{sec:transfer-data}

Two benchmarks written by other authors on reBEN images measure how well a model trained on our questions answers questions it was not trained on. Neither is used for training.

\paragraph{RSVQAxBEN.} RSVQAxBEN~\citep{rsvqaxben} contains questions generated from the original BigEarthNet labels. We keep all 29{,}501 presence questions whose image belongs to our reBEN test subset (1{,}437 images) and ask them verbatim as \texttt{noul} questions.
\label{sec:rsvqa-data}

\paragraph{BigEarthNet.txt.} The BigEarthNet.txt benchmark split~\citep{bentxt} is manually verified and contains 6{,}927 binary and 5{,}550 four-option multiple-choice questions on 1{,}065 images of the reBEN test split, on adjacency, area, count, presence, relative position, climate zone, country and season. The binary questions become \texttt{noul} questions and the multiple-choice questions become \texttt{choice} questions whose options are the a)--d) options of the original text. Many of these question kinds (adjacency, counting, relative position, season) do not occur in our training questions.
\label{sec:bentxt-data}

\subsection{EuroSAT-MS (ablation only)}
\label{sec:eurosat-data}

EuroSAT-MS~\citep{eurosat} contains 27{,}000 Sentinel-2 L1C images of $64\times64$ pixels in 10 land-cover classes. We drop the cirrus band B10 and reorder the other 12 bands as in reBEN; the files carry no band names, so we inferred the order from the data (the cirrus band has a mean of about 12 digital numbers, the 60\,m bands are the smoothest, and the band that correlates 0.98 with B08 is B8A) and check it on every conversion. We use the spatial (longitude) split of EuroSAT-Embed~\citep{geopool} and ask, per image, the land-cover class (\texttt{choice}, 10 options) and four \texttt{noul} questions on class groups (water, natural vegetation, agriculture, built-up). These 81{,}000 training questions enter only the ablation that adds a second collection (\cref{sec:ablations}); EuroSAT-MS is an L1C (top-of-atmosphere) product, unlike reBEN.

\subsection{Protocol}
\label{sec:protocol}

All models are trained on the 70{,}890 CORINE training questions and evaluated on the CORINE test set (seen and unseen templates), RSVQAxBEN and BigEarthNet.txt. Validation questions are used only for temperatures, checkpoint selection where a recipe has one, and the learning-rate pilot of \cref{sec:msi}. All benchmark images are disjoint from the training images. Hypotheses, statistics and the claim rule were fixed in a written analysis plan before any model of this comparison was trained (\cref{sec:setup}).

\section{Experiments}
\label{sec:experiments}

\subsection{Setup}
\label{sec:setup}

\paragraph{Models.} Every model is trained on the 70{,}890 CORINE training questions of \cref{sec:protocol} with seeds 0, 1 and 2, and evaluated on the same test sets.
\begin{itemize}
\item \textbf{\ej}: \cref{sec:method}; Qwen3.5-0.8B with the 12-band input layer and the official OneJev 0.8B recipe.
\item \textbf{Same backbone, generative}: Qwen3.5-0.8B with the same widened and trained input layer, the same 12-band images, state and data, fully fine-tuned to \emph{generate} the answer (next-token cross-entropy on the option text, learning rate $5\times10^{-6}$, one epoch, options permuted). At test time the answer is decoded greedily and matched to an option. This model isolates the read-out: everything else is the same as \ej.
\item \textbf{Multimodal LLMs}: Qwen3-VL-2B~\citep{qwen3vl}, InternVL3-2B~\citep{internvl3} and the remote-sensing model EarthDial-4B (RGB checkpoint)~\citep{earthdial}, fine-tuned with LoRA~\citep{lora} (rank 32, $\alpha=64$, learning rate $10^{-4}$, one epoch) to generate the answer. They accept three channels and see the true-colour composite.
\item \textbf{Fixed-vocabulary VQA (reference)}: RSVQA-style models~\citep{rsvqa} with a ResNet-50~\citep{resnet}, Swin-T~\citep{swin} or MambaVision-T~\citep{mambavision} image encoder on the 12 bands and a frozen sentence encoder~\citep{sbertmulti}, fused by element-wise product and classified over the answer vocabulary of the training questions, restricted to each question's options; 20 epochs, epoch chosen on validation. They are the standard models of remote-sensing VQA and the cheapest way to answer our questions, but they do a narrower task: they cannot answer with an option that was not in their training vocabulary. We report them as a reference for cost, not as competitors.
\item \textbf{Reference}: a U-Net~\citep{unet} trained on the whole reBEN training split segments the patch into CORINE level-2 classes, and the question generator's own rules answer from the predicted map. It is not a competitor: it is built for exactly these question templates, cannot answer other questions and gives no probabilities.
\end{itemize}

\paragraph{Analysis plan.} Hypotheses, statistics and the rule for claiming a result were written down and fixed, with a hash, before any model of this comparison was trained; later changes are dated amendments and are stated where they apply. Three hypotheses are tested on the CORINE test set:
\textbf{H1} (equivalence): the accuracy of \ej{} is within $\pm1$ point of each multimodal LLM and of the generative model with the same backbone (two one-sided tests, Holm correction over the comparisons);
\textbf{H2} (superiority): \ej{} is more accurate than each fixed-vocabulary model (one-sided, Holm);
\textbf{H3} (band use): \ej{} is more accurate than the same models with the nine non-RGB bands replaced by their training mean at test time (one-sided).
Intervals come from a paired bootstrap over test images (10{,}000 replicates; all questions of an image are resampled together) on seed-averaged predictions. A hypothesis is supported only if its test passes on the seed-averaged predictions \emph{and} in at least two of the three seeds. All other analyses are descriptive.

\paragraph{Latency.} All models are timed in one job on one H100 PCIe GPU in bfloat16, on the same 100 test images after 10 warm-up images. Each timer runs from the decoded image in memory to the answer, so image preprocessing and tokenisation are included for every model. Two scenarios are measured: one question per request (batch 1), and all the questions of an image in one request, which \ej{} answers with OneJev's engine by encoding the image once and evaluating the questions as branches of a shared prefix, the generative models answer with one batched generation and the fixed-vocabulary models with one batched forward pass.

\subsection{Speed and quality}
\label{sec:speed}

\begin{table}[t]
  \centering
  \caption{Speed against quality. CORINE: seed mean (standard deviation as subscript) of the accuracy on the 36{,}028 test questions, and mean of the per-seed ECE (15 bins, before temperature scaling). Latency per question at batch 1 and how many times slower than \ej{} each model is, from \cref{tab:latency}. Fixed-vocabulary models answer only the questions and options seen in training.}
  \label{tab:speed-accuracy}
  \small
  \IfFileExists{\paperroot tables/speed_accuracy}{\resizebox{\linewidth}{!}{\begin{tabular}{@{}llrrrrr@{}}
\toprule
 & & \multicolumn{2}{c}{CORINE test (36{,}028 q.)} & RSVQAxBEN & \multicolumn{2}{c}{Latency per question} \\
\cmidrule(lr){3-4}\cmidrule(lr){6-7}
Model & Params & Acc. (\%) & ECE & Acc. (\%) & p50 (ms) & Slower \\
\midrule
\multicolumn{7}{@{}l}{\textit{Ours}} \\
\quad \textbf{\ej{} (ours)} & \textbf{0.8B} & \textbf{86.0$_{\pm 0.1}$} & \textbf{0.006} & \textbf{81.0$_{\pm 0.6}$} & \textbf{19.9} & \textbf{--} \\
\addlinespace
\multicolumn{7}{@{}l}{\textit{Same backbone and input, generative fine-tuning}} \\
\quad Qwen3.5-0.8B, generative & 0.8B & 86.8$_{\pm 0.2}$ & 0.005 & 80.5$_{\pm 0.6}$ & 134.7 & 6.8$\times$ \\
\addlinespace
\multicolumn{7}{@{}l}{\textit{Multimodal LLMs, LoRA fine-tuning}} \\
\quad Qwen3-VL-2B & 2.1B & 86.6$_{\pm 0.1}$ & 0.005 & 80.6$_{\pm 0.1}$ & 111.6 & 5.6$\times$ \\
\quad InternVL3-2B & 2.1B & 87.1$_{\pm 0.0}$ & 0.006 & 81.2$_{\pm 0.4}$ & 84.5 & 4.3$\times$ \\
\quad EarthDial-4B & 4.2B & 89.5$_{\pm 0.1}$ & 0.006 & 82.4$_{\pm 0.4}$ & 171.8 & 8.7$\times$ \\
\addlinespace
\multicolumn{7}{@{}l}{\textit{Reference: fixed-vocabulary VQA}} \\
\quad ResNet-50 (CNN) & 0.14B & 87.5$_{\pm 0.2}$ & 0.089 & 79.4$_{\pm 0.1}$ & 12.1 & 0.6$\times$ \\
\quad Swin-T (Transformer) & 0.15B & 86.5$_{\pm 0.1}$ & 0.047 & 79.6$_{\pm 0.3}$ & 16.2 & 0.8$\times$ \\
\quad MambaVision-T (Mamba) & 0.15B & 85.9$_{\pm 0.3}$ & 0.064 & 78.4$_{\pm 0.2}$ & 15.3 & 0.8$\times$ \\
\bottomrule
\end{tabular}
}}{\todo{table generated when all models are evaluated}}
\end{table}

\begin{table*}[t]
  \centering
  \caption{Inference cost of every model, measured in one job on one H100 PCIe GPU (bfloat16; 100 test images after 10 warm-up images; each timer runs from the decoded image in memory to the answer, preprocessing and tokenisation included). \emph{Slower}: p50 latency of the model divided by that of \ej{} in the same scenario (below 1: faster than \ej). \emph{Image}: p50 latency of a request with all the questions of an image (6.1 on average); \emph{Per q.}: mean latency per question in that scenario. In the second scenario \ej{} encodes the image once and answers all its questions as branches of a shared prefix; the generative models answer them in one batched generation and the fixed-vocabulary models in one batched forward pass.}
  \label{tab:latency}
  \small\setlength{\tabcolsep}{4pt}
  \IfFileExists{\paperroot tables/latency}{\resizebox{\linewidth}{!}{\begin{tabular}{@{}lllrrrrrrr@{}}
\toprule
 & & & \multicolumn{3}{c}{One question per request} & \multicolumn{3}{c}{All questions of an image} & \\
\cmidrule(lr){4-6}\cmidrule(lr){7-9}
Model & Params & Input & p50 (ms) & p95 (ms) & Slower & Image (ms) & Per q. (ms) & Slower & Mem. (GB) \\
\midrule
\multicolumn{10}{@{}l}{\textit{Ours}} \\
\quad \textbf{\ej{} (ours)} & \textbf{0.8B} & \textbf{12 bands} & \textbf{19.9} & \textbf{20.1} & \textbf{--} & \textbf{42.8} & \textbf{7.9} & \textbf{--} & \textbf{3.5} \\
\addlinespace
\multicolumn{10}{@{}l}{\textit{Same backbone and input, generative fine-tuning}} \\
\quad Qwen3.5-0.8B, generative & 0.8B & 12 bands & 134.7 & 144.4 & 6.8$\times$ & 191.9 & 36.9 & 4.5$\times$ & 2.2 \\
\addlinespace
\multicolumn{10}{@{}l}{\textit{Multimodal LLMs, LoRA fine-tuning}} \\
\quad Qwen3-VL-2B & 2.1B & RGB & 111.6 & 146.7 & 5.6$\times$ & 176.2 & 34.5 & 4.1$\times$ & 4.6 \\
\quad InternVL3-2B & 2.1B & RGB & 84.5 & 113.0 & 4.3$\times$ & 157.3 & 30.1 & 3.7$\times$ & 4.6 \\
\quad EarthDial-4B & 4.2B & RGB & 171.8 & 217.3 & 8.7$\times$ & 431.6 & 74.9 & 10.1$\times$ & 13.5 \\
\addlinespace
\multicolumn{10}{@{}l}{\textit{Reference: fixed-vocabulary VQA}} \\
\quad ResNet-50 (CNN) & 0.14B & 12 bands & 12.1 & 13.8 & 0.6$\times$ & 14.9 & 4.4 & 0.3$\times$ & 0.7 \\
\quad Swin-T (Transformer) & 0.15B & 12 bands & 16.2 & 16.4 & 0.8$\times$ & 17.9 & 4.3 & 0.4$\times$ & 0.8 \\
\quad MambaVision-T (Mamba) & 0.15B & 12 bands & 15.3 & 15.9 & 0.8$\times$ & 18.6 & 4.8 & 0.4$\times$ & 0.8 \\
\bottomrule
\end{tabular}
}}{\todo{latency table: measurement running}}
\end{table*}

\Cref{tab:latency} reports the cost of every model. With one question per request, \ej{} answers in 19.9\,ms. The multimodal LLMs need 84.5--171.8\,ms: \textbf{\ej{} is 4.3 to 8.7 times faster}, with 2.6 to 5 times fewer parameters and less GPU memory (3.5\,GB against 4.6--13.5\,GB). The cleanest comparison is with the same backbone fine-tuned generatively: same weights, same 12-band images, same questions; the only difference is that \ej{} reads the option letters at one position instead of decoding the answer, and this makes it \textbf{6.8 times faster} (134.7\,ms).

When all the questions about an image are asked together, every model benefits, but not equally. \ej{} encodes the image once and answers the 6.1 questions of an average test image in 42.8\,ms (7.9\,ms per question). The generative models, which answer the same questions in one batched generation, need 157.3--431.6\,ms per image: \textbf{\ej{} is 3.7 to 10.1 times faster per image}, and 4.5 times faster than its own backbone used generatively. The fixed-vocabulary models are faster in both scenarios (12.1--16.2\,ms per question, 14.9--18.6\,ms per image), as expected from their size (0.14--0.15B parameters); they set the cost of a model that answers only the questions and options of its training vocabulary. With one question per request \ej{} takes 1.2--1.6 times as long while answering any declared option set.

Latency depends on the load of the node: in a first measurement job, with the second scenario for \ej{} only, the generative models were up to 26\% slower with one question per request (e.g.\ 141.0 against 111.6\,ms for Qwen3-VL-2B), while \ej{} took 19.9\,ms in both. \Cref{tab:latency} reports the second job, the only one with every model in both scenarios and the less favourable to \ej.

\Cref{tab:speed-accuracy} puts the cost next to the quality. The speed does not come from a much less accurate model: on CORINE, \ej{} is 0.6 points below Qwen3-VL-2B and 1.2 points below InternVL3-2B, which are 5.6 and 4.3 times slower, and as well calibrated (ECE 0.006 against 0.005--0.006). EarthDial-4B, a remote-sensing model 5 times larger and 8.7 times slower, is 3.5 points more accurate.

\subsection{Accuracy, calibration and transfer}
\label{sec:results}

\begin{table*}[t]
  \centering
  \caption{All models trained on the same 70{,}890 CORINE questions (three seeds). CORINE: accuracy (\%) overall (seed mean, standard deviation as subscript) and per question type, seed-averaged; ECE: mean of the per-seed ECEs, 15 bins, before temperature scaling. RSVQAxBEN: 29{,}501 presence questions. BigEarthNet.txt: 6{,}927 binary and 5{,}550 four-option questions; the fixed-vocabulary models cannot answer them (their answer vocabulary does not cover them). Transfer tests are never used for training.}
  \label{tab:main}
  \small\setlength{\tabcolsep}{3.6pt}
  \resizebox{\linewidth}{!}{\begin{tabular}{@{}lllrrrrrrrrrr@{}}
\toprule
 & & & \multicolumn{7}{c}{CORINE test (36{,}028 questions)} & RSVQA- & \multicolumn{2}{c}{BigEarthNet.txt} \\
\cmidrule(lr){4-10}\cmidrule(lr){12-13}
Model & Params & Input & All & Dom. & Pres. & Share & Loc. & Comp. & ECE & xBEN & Binary & MC \\
\midrule
\multicolumn{13}{@{}l}{\textit{Ours}} \\
\quad \textbf{\ej{} (ours)} & \textbf{0.8B} & \textbf{12 bands} & \textbf{86.0$_{\pm 0.1}$} & \textbf{83.6} & \textbf{93.2} & \textbf{82.9} & \textbf{66.1} & \textbf{87.9} & \textbf{0.006} & \textbf{81.0} & \textbf{59.6} & \textbf{40.1} \\
\addlinespace
\multicolumn{13}{@{}l}{\textit{Same backbone and input, generative fine-tuning}} \\
\quad Qwen3.5-0.8B, generative & 0.8B & 12 bands & 86.8$_{\pm 0.2}$ & 85.0 & 93.6 & 83.7 & 66.6 & 88.5 & 0.005 & 80.5 & 59.3 & 17.1 \\
\addlinespace
\multicolumn{13}{@{}l}{\textit{Multimodal LLMs, LoRA fine-tuning}} \\
\quad Qwen3-VL-2B & 2.1B & RGB & 86.6$_{\pm 0.1}$ & 84.6 & 93.6 & 83.9 & 66.5 & 87.9 & 0.005 & 80.6 & 60.4 & 39.6 \\
\quad InternVL3-2B & 2.1B & RGB & 87.1$_{\pm 0.0}$ & 85.4 & 93.9 & 84.0 & 66.4 & 89.3 & 0.006 & 81.2 & 59.9 & 38.9 \\
\quad EarthDial-4B & 4.2B & RGB & 89.5$_{\pm 0.1}$ & 88.9 & 94.9 & 86.5 & 70.7 & 91.1 & 0.006 & 82.4 & 62.0 & 42.9 \\
\addlinespace
\multicolumn{13}{@{}l}{\textit{Reference: fixed-vocabulary VQA}} \\
\quad ResNet-50 (CNN) & 0.14B & 12 bands & 87.5$_{\pm 0.2}$ & 87.8 & 94.2 & 84.0 & 56.2 & 90.4 & 0.089 & 79.4 & -- & -- \\
\quad Swin-T (Transformer) & 0.15B & 12 bands & 86.5$_{\pm 0.1}$ & 88.8 & 94.7 & 86.5 & 27.7 & 91.5 & 0.047 & 79.6 & -- & -- \\
\quad MambaVision-T (Mamba) & 0.15B & 12 bands & 85.9$_{\pm 0.3}$ & 87.9 & 94.3 & 85.2 & 29.7 & 90.5 & 0.064 & 78.4 & -- & -- \\
\addlinespace
\multicolumn{13}{@{}l}{\textit{Reference: segmentation + the generator's rules}} \\
\quad U-Net + rules (reference) & 0.03B & 12 bands & 91.0 & 91.5 & 94.6 & 89.1 & 74.8 & 92.0 & -- & -- & -- & -- \\
\bottomrule
\end{tabular}
}
\end{table*}

\paragraph{Hypotheses.} \textbf{H3 is supported}: replacing the nine non-RGB bands with their training mean lowers the CORINE accuracy of \ej{} by 4.5 points (95\% interval 4.2--4.9; all three seeds), and by 2.3 and 1.9 points on RSVQAxBEN and BigEarthNet.txt. The model uses the bands it was given, although their kernels grow to only about 1\% of the norm of the RGB kernels in one epoch at the recipe's learning rate. \textbf{H1 is not supported.} Against Qwen3-VL-2B the seed-averaged difference ($-0.6$ points, 90\% interval $-0.9$ to $-0.4$) lies inside the $\pm1$ margin, but only one seed of three passes, so the claim rule rejects it; InternVL3-2B is 1.2 points and EarthDial-4B 3.5 points more accurate. Against its own backbone fine-tuned to generate the answer, \ej{} is 0.8 points less accurate (90\% interval $-1.0$ to $-0.6$): the interval touches the margin and only one seed passes, so equivalence is not supported either, although the difference is small. \textbf{H2 is not supported}: on the questions and options they were trained on, the fixed-vocabulary ResNet-50 is 1.5 points and Swin-T 0.5 points more accurate than \ej, and MambaVision-T is equal ($+0.1$, interval $-0.3$ to $+0.4$).

\paragraph{Where the accuracy differs.} \Cref{tab:main} shows that the models do not fail on the same questions. The fixed-vocabulary models are best on dominant-class questions (87.8--88.8\% against 83.6\% for \ej) and weakest on location questions: Swin-T and MambaVision-T answer which quadrant holds a class barely above chance (27.7\% and 29.7\%; four options, chance 25\%), ResNet-50 at 56.2\%, while \ej{} reaches 66.1\% and the multimodal LLMs 66.4--70.7\%. Rephrasing the questions with templates never seen in training does not change the accuracy of \ej{} (86.0\% on both sets).

\paragraph{Calibration.} The probabilities of \ej{} are calibrated without temperature scaling (ECE 0.006). The fixed-vocabulary models, also trained with a proper scoring rule, have an ECE of 0.047--0.089. The LoRA-tuned multimodal LLMs are as well calibrated as \ej{} when the probability of their generated answer is read out (0.005--0.006); \ej{} obtains the same calibration without decoding.

\paragraph{Transfer to questions written by others.} On RSVQAxBEN, whose presence questions were generated by other authors from the original BigEarthNet labels, \ej{} reaches 81.0\%, within the range of the multimodal LLMs (80.6--82.4\%); the fixed-vocabulary models reach 78.4--79.6\%. On BigEarthNet.txt every model trained on our questions stays well below its training accuracy: about 60\% on binary questions (chance: 50\%) and about 40\% on four-option questions (chance: 25\%), with EarthDial-4B 1.6--2.4 points above the others on binary and 2.8--4.0 points on four-option questions. Here the read-out matters: the options of these questions were never seen in training, and the same backbone fine-tuned to generate answers with a text that matches none of the four options for 45.9\% of the multiple-choice questions (for example a CORINE class name from our training questions), which leaves it at 17.1\%, below chance; the multimodal LLMs do so for 1.0--5.9\%. \ej{} cannot: it returns a distribution over the declared options by construction. Many of its questions concern adjacency, counts, relative positions or seasons, which our training questions never ask; these are failures of transfer shared by all models, not of the read-out.

\paragraph{Reference.} Answering CORINE questions with the generator's own rules on a U-Net map gives 91.0\%. The information needed is therefore in the 12 bands; the gap to the models that read the image and the question is the cost of answering arbitrary questions instead of the generator's templates.

\subsection{Ablations}
\label{sec:ablations}

\begin{table}[t]
  \centering
  \caption{One-factor ablations of \ej{} (seed 0). $\Delta$: accuracy of the variant minus that of \ej, in points, with the 95\% paired bootstrap interval; positive values mean that the variant is more accurate. Pre-specified and descriptive: no correction for multiple comparisons and no claim rule.}
  \label{tab:ablation}
  \small\setlength{\tabcolsep}{3.5pt}
  \resizebox{\linewidth}{!}{\begin{tabular}{@{}lrlrlrl@{}}
\toprule
 & \multicolumn{2}{c}{CORINE} & \multicolumn{2}{c}{RSVQAxBEN} & \multicolumn{2}{c}{BigEarthNet.txt} \\
\cmidrule(lr){2-3}\cmidrule(lr){4-5}\cmidrule(lr){6-7}
Variant & Acc. & $\Delta$ [95\% CI] & Acc. & $\Delta$ [95\% CI] & Acc. & $\Delta$ [95\% CI] \\
\midrule
\ej{} (seed 0) & \textbf{85.9} & -- & \textbf{80.3} & -- & \textbf{50.8} & -- \\
\midrule
whole vision encoder trained & 88.4 & +2.5 [+2.2, +2.8] & 83.0 & +2.7 [+2.2, +3.1] & 51.1 & +0.3 [-0.4, +1.1] \\
new channels start from the mean RGB kernel & 84.3 & -1.6 [-2.0, -1.3] & 80.3 & +0.0 [-0.4, +0.5] & 47.5 & -3.3 [-4.1, -2.5] \\
acquisition date and country in the state & 86.4 & +0.5 [+0.1, +0.8] & 81.9 & +1.5 [+1.2, +1.9] & 50.7 & -0.1 [-0.9, +0.6] \\
+ 81{,}000 EuroSAT-MS questions & 86.5 & +0.6 [+0.3, +0.9] & 78.9 & -1.5 [-1.9, -1.1] & 49.2 & -1.5 [-2.3, -0.8] \\
50\% of the training patches & 84.5 & -1.4 [-1.7, -1.1] & 80.9 & +0.5 [+0.2, +0.9] & 50.1 & -0.6 [-1.4, +0.1] \\
25\% of the training patches & 83.2 & -2.7 [-3.1, -2.4] & 79.3 & -1.1 [-1.5, -0.7] & 49.3 & -1.5 [-2.3, -0.7] \\
cross-entropy only (no Brier term) & 86.2 & +0.2 [-0.0, +0.5] & 82.1 & +1.7 [+1.3, +2.2] & 50.7 & -0.1 [-0.7, +0.6] \\
no state augmentation & 85.7 & -0.2 [-0.5, +0.0] & 81.9 & +1.6 [+1.2, +2.0] & 51.0 & +0.2 [-0.5, +0.9] \\
\bottomrule
\end{tabular}
}
\end{table}

\Cref{tab:ablation} varies one factor at a time.
\begin{itemize}
\item \textbf{The frozen vision encoder limits accuracy.} Training the whole vision encoder, not only its input layer, raises CORINE accuracy by 2.5 points to 88.4\% and RSVQAxBEN accuracy by 2.7 points to 83.0\%, at the same calibration (ECE 0.006) and the same inference cost. With one seed this is an indication, not a tested result; the OneJev recipe, which we follow, keeps the encoder frozen.
\item \textbf{Zero initialisation of the new bands is better} than starting them from the mean of the RGB kernels ($-1.6$ points on CORINE, $-3.3$ on BigEarthNet.txt): with zero kernels the model starts exactly from the RGB model.
\item \textbf{More data helps}: with 50\% and 25\% of the training patches the CORINE accuracy drops by 1.4 and 2.7 points.
\item \textbf{Acquisition date and country} in the state add 0.5 points on CORINE and 1.5 on RSVQAxBEN. We leave them out by design: the model must work from the image alone, and other collections (e.g.\ EuroSAT) carry no date.
\item \textbf{A second collection} (81{,}000 EuroSAT-MS questions, L1C product) adds 0.6 points on CORINE but costs 1.5 points on RSVQAxBEN; the same model answers the EuroSAT-MS land-cover question for 93.5\% of the test images of the spatial split.
\item \textbf{Loss and augmentation.} Without the Brier term, or without the serialisation augmentation of the state, CORINE accuracy is unchanged within 0.2 points, while RSVQAxBEN accuracy is 1.6--1.7 points higher. These recipe components do not help transfer in our setting.
\end{itemize}

\section{Limitations}
\label{sec:limitations}

\paragraph{Closed answers only.} \ej{} answers questions with declared options. Open requests (``describe this image'') are outside the paradigm by construction, and a user question has to be mapped to a typed question with options. In our benchmarks this mapping is given; in an application it requires rules or a separate language model that only parses the question.

\paragraph{Accuracy.} The pre-specified hypotheses of equivalence with the multimodal language models and of superiority over the fixed-vocabulary models are not supported. \ej{} is 0.6--1.2 points less accurate than the 2B multimodal models and 3.5 points less than EarthDial-4B, and on the questions it was trained on, the fixed-vocabulary ResNet-50 is both more accurate and faster. The case for \ej{} rests on speed relative to the multimodal models, at a small cost in accuracy, with calibrated answers to any declared option set.

\paragraph{Frozen vision encoder.} We keep the OneJev recipe, which freezes the vision encoder; only its new input layer is trained. The ablation that trains the whole encoder gains 2.5 points with one seed. This, and larger backbones, need a pre-specified test with several seeds before any claim.

\paragraph{Latency.} Latency depends on the inference software and on the load of the node. \ej{} runs on the OneJev engine; the generative models run \texttt{generate} from Transformers without an optimised serving engine, which would narrow the gap. Between two measurement jobs the generative models varied by up to 26\% while \ej{} did not; we report the job less favourable to \ej. The fixed cost of one forward pass against one pass per generated token does not depend on the engine.

\paragraph{Inputs of the baselines.} The multimodal language models accept three channels and see the true-colour composite, so they receive less spectral information than \ej{} and the fixed-vocabulary models. This is how these models are used, but it means that part of the difference in accuracy, in either direction, may come from the input rather than the model.

\paragraph{Scale and coverage of the data.} We train on 5\% of reBEN, 11{,}893 training images from ten European countries in 2017--2018; the ablations show that accuracy still grows with data. Calibration across regions and seasons outside the training distribution is untested.

\paragraph{Labels and questions.} CORINE has a minimum mapping unit of 25\,ha, so the reference masks are approximate at 10\,m; our questions inherit this approximation and are generated from templates. The held-out templates show that the model is not tied to the training phrasings, but they do not cover the variety of real user questions, and on BigEarthNet.txt, whose question types differ from ours, every model transfers poorly.

\paragraph{Contamination.} EarthDial's training data include BigEarthNet, and 3{,}951 of our 5{,}991 test images belong to the RSVQAxBEN training split, so EarthDial may have seen our test images; this would favour it. The pre-training data of the Qwen and InternVL backbones are not disclosed.

\section{Conclusion}
\label{sec:conclusion}

\ej{} shows that a vision-language model can read all 12 Sentinel-2 bands through a widened input layer and answer typed questions about them without generating text. Reading the option letters at one position makes it 6.8 times faster than the same backbone fine-tuned to generate the answer, and 4.3 to 8.7 times faster than multimodal language models 2.6 to 5 times larger, at an accuracy 0.6--1.2 points below the 2B models; its answers are always valid and its probabilities calibrated without decoding. With one question per request it takes 1.2--1.6 times as long as VQA models that answer only from a fixed vocabulary, while accepting any declared option set. The extra bands are used, and the frozen vision encoder is the main limit on accuracy.

Next steps are training the vision encoder and larger backbones under a pre-specified plan, training on the whole of BigEarthNet v2.0, and testing outside the training distribution. Because each answer costs milliseconds and no text, the model also suits monitoring: with pairs or time series of Sentinel-2 images in the state, an agent could ask change questions over large areas and many dates at a cost that generative models do not allow.

\paragraph{Code and model.} Code: \url{https://github.com/CosmoIknosLab/EarthJev}. Model: \url{https://huggingface.co/dBenf/EarthJev}.

\section*{Acknowledgements}
This study was partially supported by the PNRR MUR project PE0000013-FAIR.

\bibliographystyle{plainnat}
\bibliography{\paperroot refs}

\appendix
\section{Example state and question}
\label{app:state}

\Cref{lst:state} shows the state of a CORINE test question as serialised at inference; the image placeholder is replaced by the 64 visual tokens of the 12-band patch. The state is the same for every patch: it describes the sensor and contains nothing derived from the image. The models that accept only RGB receive the same state with the sensor described as a true-colour composite (B4, B3, B2).

\begin{lstlisting}[caption={State of a CORINE test question (reBEN patch \texttt{S2A\_MSIL2A\_20170613T101031\_N9999\_R022\_T33UUP\_29\_59}).},label={lst:state}]
{
  "sensor": "Sentinel-2 L2A (surface reflectance), multispectral",
  "bands": "12 bands: B01, B02, B03, B04, B05, B06, B07, B08, B8A, B09, B11, B12
            (20 m and 60 m bands resampled to 10 m)",
  "ground_sample_distance": "10 m",
  "patch": "120 x 120 pixels (1.2 km x 1.2 km), north up",
  "image": "<image:1>"
}
Question (choice): Which CORINE land-cover class occupies the largest area here?
Options: Artificial surfaces. | Agricultural areas. | Forest and seminatural areas. | Wetlands. | Water bodies.
\end{lstlisting}

\section{Question templates}
\label{app:templates}

\Cref{tab:templates} lists the question templates. \{c\}, \{a\}, \{b\} are CORINE class names at level 1 or 2.

\begin{table}[h]
  \centering
  \caption{Question templates. Seen templates are used for training, validation and the main test set; the unseen template of each type is used only for the second rendering of the test set.}
  \label{tab:templates}
  \small
  \begin{tabular}{@{}lp{0.62\linewidth}l@{}}
    \toprule
    Type & Template & Split \\
    \midrule
    dominant & Which land-cover class covers most of this image? & seen \\
             & What is the dominant land cover in this Sentinel-2 patch? & seen \\
             & Which CORINE land-cover class occupies the largest area here? & seen \\
             & Looking at this satellite image, which kind of land covers the biggest part of it? & unseen \\
    \midrule
    presence & This image contains \{c\}. & seen \\
             & Is there any \{c\} in this patch? & seen \\
             & Does \{c\} appear in this satellite image? & seen \\
             & Can you see \{c\} anywhere in this picture? & unseen \\
    \midrule
    share    & What share of the image is covered by \{c\}? & seen \\
             & How much of this patch is \{c\}? & seen \\
             & Which fraction of the area is occupied by \{c\}? & seen \\
             & Roughly what percentage of this scene is \{c\}? & unseen \\
    \midrule
    location & In which part of the image is most of the \{c\}? & seen \\
             & Where in this patch is the \{c\} mainly located? & seen \\
             & Which quadrant contains the largest part of the \{c\}? & seen \\
             & If you split the image into four, where would you find most of the \{c\}? & unseen \\
    \midrule
    compare  & Which covers more of the image: \{a\} or \{b\}? & seen \\
             & Is there more \{a\} or more \{b\} in this patch? & seen \\
             & Which class has the larger area here, \{a\} or \{b\}? & seen \\
             & Comparing \{a\} and \{b\}, which one takes up more space in this scene? & unseen \\
    \bottomrule
  \end{tabular}
\end{table}

\end{document}